\documentclass[lettersize,journal]{IEEEtran}
\usepackage{cite}
\usepackage[ruled]{algorithm2e}
\usepackage{amsmath,amssymb,amsfonts}
\usepackage{algorithmic}
\usepackage{array}
\usepackage{balance}
\usepackage{cellspace}
\usepackage{graphicx}
\usepackage{hhline}
\usepackage{makecell}
\usepackage{multirow}
\usepackage{textcomp}
\usepackage{soul}
\SetKwInput{KwFeature}{Input}
\SetKwInput{KwOutput}{Output}

\begin{document}

\title{LingML: Linguistic-Informed Machine Learning for Enhanced Fake News Detection}

\author{Jasraj Singh,~Fang Liu,~\IEEEmembership{Member,~IEEE},~Hong Xu,~Bee Chin Ng,~and~Wei Zhang,~\IEEEmembership{Member,~IEEE}
\thanks{This work was supported in part by the School of Social Sciences at Nanyang Technological University, A*STAR under its MTC Programmatic (Award M23L9b0052), SIT’s Ignition Grant (STEM) (Grant ID: IG (S) 2/2023 – 792), and the Ministry of Education, Singapore, under the Academic Research Tier 1 Grant (Grant ID: GMS 693). (\textit{Corresponding author: Fang Liu})}
\thanks{Jasraj Singh is with the Nanyang Technological University (NTU), Singapore, and the University College London (UCL), United Kingdom (e-mail: jasraj001@e.ntu.edu.sg and jasraj.singh.23@ucl.ac.uk).}
\thanks{Fang Liu is with the School of Science and Technology, Singapore University of Social Sciences, Singapore 599494 (e-mail: liufang@suss.edu.sg).}
\thanks{Hong Xu is with the School of Social Sciences at Nanyang Technological University, Singapore 639798 (e-mail: xuhong@ntu.edu.sg).}
\thanks{Bee Chin Ng is with the School of Humanities at Nanyang Technological University, Singapore 639798 (e-mail: mbcng@ntu.edu.sg).}
\thanks{Wei Zhang is with the Information and Communications Technology Cluster, Singapore Institute of Technology, Singapore 138683 (e-mail: wei.zhang@singaporetech.edu.sg).}
\thanks{Manuscript received January 1, 2024; revised January 1, 2024.}}

\markboth{Journal of \LaTeX\ Class Files,~Vol.~1, No.~1, January~2024}%
{Shell \MakeLowercase{\textit{et al.}}: A Sample Article Using IEEEtran.cls for IEEE Journals}

\IEEEpubid{0000--0000/00\$00.00~\copyright~2021 IEEE}

\maketitle

\bstctlcite{IEEEexample:BSTcontrol}

\maketitle
\begin{abstract}
Nowadays, Information spreads at an unprecedented pace in social media and discerning truth from misinformation and fake news has become an acute societal challenge. Machine learning (ML) models have been employed to identify fake news but are far from perfect with challenging problems like limited accuracy, interpretability, and generalizability. In this paper, we enhance ML-based solutions with linguistics input and we propose \texttt{LingML}, linguistic-informed ML, for fake news detection. We conducted an experimental study with a popular dataset on fake news during the pandemic. The experiment results show that our proposed solution is highly effective. There are fewer than two errors out of every ten attempts with only linguistic input used in ML and the knowledge is highly explainable. When linguistics input is integrated with advanced large-scale ML models for natural language processing, our solution outperforms existing ones with 1.8\% average error rate. \texttt{LingML} creates a new path with linguistics to push the frontier of effective and efficient fake news detection. It also sheds light on real-world multi-disciplinary applications requiring both ML and domain expertise to achieve optimal performance. 
\end{abstract}
\section{Introduction}

\IEEEPARstart{T}{he} advent of the digital age has ushered in connectivity and information sharing primarily through social media platforms. Such interconnectedness indeed has transformed the way we communicate and brought us various benefits, but it has also given rise to the spread of misinformation and fake news \cite{wu2019misinformation}. In major events like the recent pandemic, social media became a breeding ground of unverified claims, conspiracy theories, etc., which offset its contribution to disseminating timely and useful news. With events carrying high stakes and uncertainties like the pandemic, people to some extent are more susceptible to misleading narratives. The consequences can be profound, e.g., shaping public opinion and eroding trust, if the spread of fake news is not well controlled. 

Preventing the negative consequences requires fake news detection and to do this effectively and efficiently is highly challenging. With the huge volume of information that is created every day and circulated online with ever-increasing social media activities, it is nearly impossible to manually sift through all the news and identify the fake ones fast enough. There is an urgent demand to design and develop automated and accurate solutions to detect fake news and curb the dissemination of misinformation in the social media era.

\IEEEpubidadjcol
A promising technology for fake news detection is machine learning (ML) which has a great potential for fighting against the fake news proliferation on social media platforms \cite{febrinanto2023graph}. For ML, fake news detection is essentially a classification problem. It distinguishes fake and non-fake news and assigns a \texttt{true} label to the fake ones and \texttt{false} otherwise. Natural language processing (NLP), a main category of ML, has been studied and applied to detect fake news because of its strength in analyzing text-based data like news \cite{hu2022deep}. NLP models have become larger and larger to achieve improved performance and several large language models (\texttt{LLM}) have been proposed in recent years \cite{keller2023chat}. Some \texttt{LLM} have also been specialized for certain topics, e.g., CT-BERT \cite{muller2023covid} for COVID-19, and have shown encouraging results.

However, existing solutions including advanced \texttt{LLM} are not yet perfect. First, they are not accurate enough. The detection error rate has been improved with newer and more advanced models in recent years. But the improvement is becoming harder and minimally marginal with the models being more complex and suffering from diminishing returns. Complexity comes with poor interpretability. The process of analyzing news and producing detection outcomes is almost impossible to understand and the solutions are often referred to as a black-box \cite{zhang2020demystifying}. As a result, many of the existing ML or \texttt{LLM} solutions have not been deployed in real-world systems where the public and researchers, e.g., from social science backgrounds, remain distrustful and skeptical. A model, even shown to be accurate, maybe over-fitted. It means that the model is only accurate for certain types of data that are used for training the model and the error rate can increase significantly for other data \cite{liu2022transline}. Social media is changing fast with new buzzwords, acronyms, etc., and the model may not be generalized enough for the changes. Yet, a model normally continues producing detection output without necessary explanation and its confidence is not conveyed to the users. 

We argue in this paper that ML alone cannot solve the problem and we propose to leverage upon a domain knowledge which has been scientifically attested. Specifically, we design and develop linguistic-informed ML \texttt{LingML} for fake news detection. With our linguistic expertise, we extract linguistic knowledge from news, in the form of numerical features. The knowledge is scientific and explainable and reflects the differences between fake and real news. We propose two different ways to utilize such knowledge in ML, to not only further reduce the error rate but also offer improved explainability to our detection solutions. We summarize our main contributions as follows.

\begin{itemize}
    \item We developed ML models \texttt{LingM} driven by linguistic knowledge only and studied and discussed the importance of different linguistic features.
    \item We proposed \texttt{LingL} which integrates linguistic knowledge into advanced \texttt{LLM} to utilize the strength of both through \texttt{LLM}-based encoding and feature fusion. 
    \item  We conducted an experimental study and demonstrated that linguistic knowledge is vital for performance improvement for both \texttt{LingM} and \texttt{LingL}, achieving the best average fake news detection error rate of 1.8\%.
\end{itemize}  

The impact of this research is many-sided. The most direct impact is that linguistic knowledge is shown to be useful for complementing and enhancing ML-based solutions with performance improvements in different aspects. The linguistic knowledge extraction, utilization, and understanding may inspire further research for fake news detection with scientific foundations. \texttt{LingML} is also within the spectrum of the emerging physics-informed machine learning and can shed light on other realistic applications that are multi-disciplinary. 

The rest of the paper is organized as follows. The system architecture of the proposed \texttt{LingML} is presented in Section \ref{sec:system}. Section \ref{sec:method} describes linguistic knowledge and our methodology in detail and Section \ref{sec:exp} presents an experimental study of the proposed solution as well as results discussion. Finally, we conclude the paper in Section \ref{sec:conclusion}.
\section{System Architecture} 
\label{sec:system}
\begin{figure}
  \centering  \includegraphics[width=0.48\textwidth]{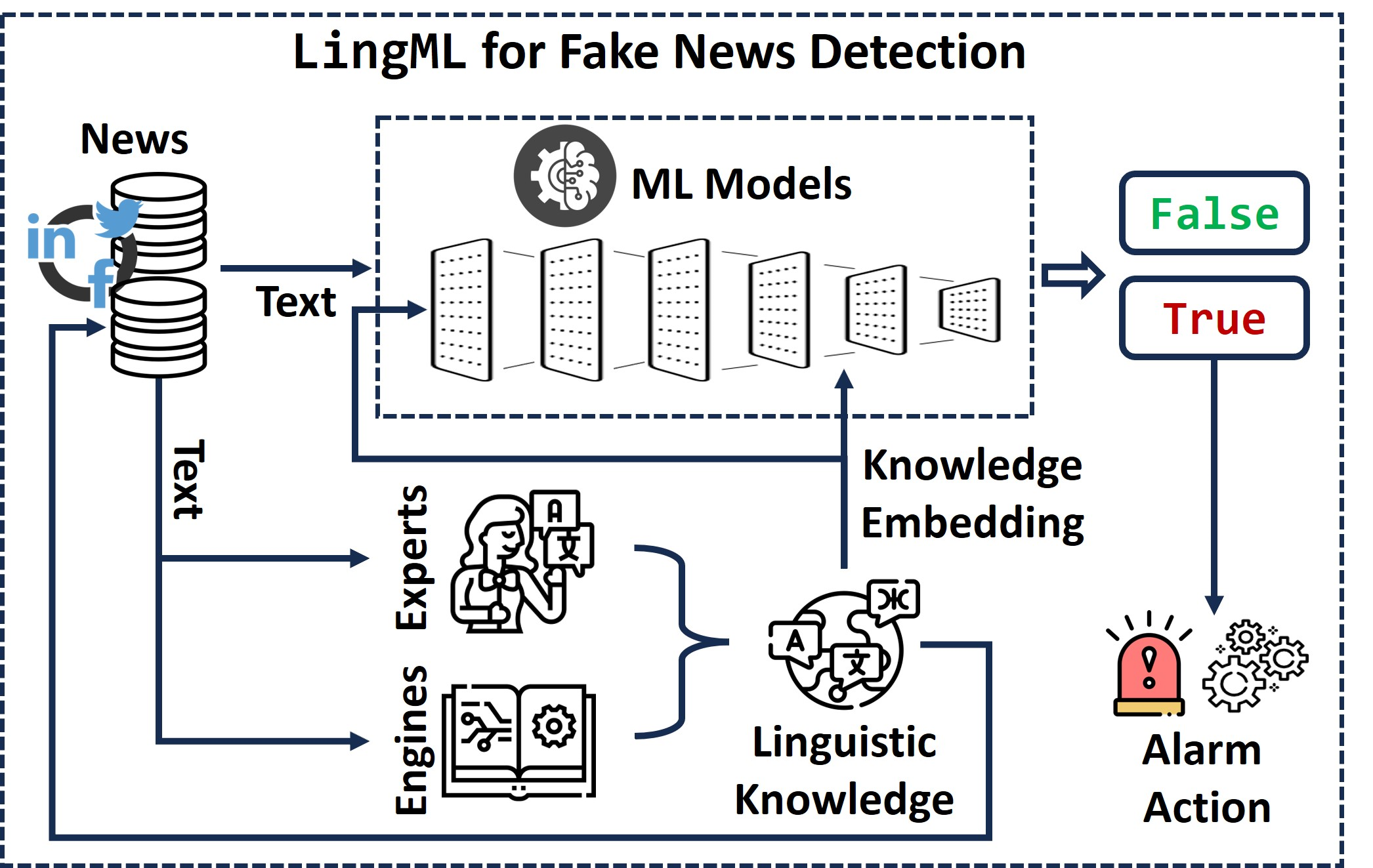}
  \caption{The system architecture of the \texttt{LingML} for fake news detection. As an ML-based system, the news is the input and the output is \texttt{true} if it is fake news and \texttt{false} otherwise. The key technical challenge in the system is knowledge representation and utilization. \texttt{LingML} extracts linguistic knowledge from the input news and produces ML digestible linguistic features. The knowledge can be integrated into \texttt{LingML} in different ways and stages for achieving the optimal detection performance and the system reacts with risk mitigation measures once a fake news is detected.}
\label{fig:sys-arch}
\end{figure}

In this section, we present the system architecture of \texttt{LingML} with ML-based detection and linguistic knowledge integration. An illustration of the system is shown in Fig. \ref{fig:sys-arch}.

\subsection{ML-based Fake News Detection}
Fake news can be detected through different ways and ML is one of the promising technologies for fast and accurate detection. In typical ML-based settings, the input of a solution is news (e.g., from social media and instant messaging) for fact-checking. The solution normally conducts data pre-processing for the input to make sure the processed news is compatible with the ML model, e.g., text-based. The ML model analyzes the news as a classification problem and outputs the detection label of the news, i.e., \texttt{true} for fake news and \texttt{false} otherwise. Once fake news is detected, follow-up system operations (e.g., double-checking, labeling, and reporting) can be triggered to mitigate potential risks.

\subsection{Linguistic Knowledge Integration}
The key novelty of this paper is the integration of linguistic knowledge in the ML-based detection models. Our idea is based on the fact that existing models generally are configured to deal with text-based information for fake news detection because the news itself is largely text-based. There are several potential ways and stages of linguistic knowledge integration. First, the knowledge needs to be represented in an ML digestible format and this can be accomplished through extracting linguistic features. Such feature extraction relies heavily on linguistic expertise and tools and the goal is to numericalize linguistic knowledge. The numerical knowledge becomes compatible with ML and can be used for model training and detection through different integration schemes.

Overall, we consider ML-based fake news detection in this paper and we propose to integrate linguistic knowledge into ML models. We expect the domain knowledge to complement and enhance data-driven ML for achieving improved system performance in terms of detection accuracy, reliability, etc. The details of our methodology are presented below.
\section{Methodology} 
\label{sec:method}
In this section, we present the detailed methodology of \texttt{LingML} including linguistic knowledge representation and two knowledge usage methods for fake news detection.

\subsection{Linguistic Knowledge and Feature Extraction} 
\label{subsec:feature-extraction}
As discussed above, linguistic knowledge is hardly used in existing ML-based solutions. A typical ML process starts with news and ends with the label of the news and such a process is mostly data-driven without linguistic expertise. Indeed, fake news detection is not only an ML problem. It has also been an active research topic in the linguistics research community. 

Researchers working on the language of fake news detection analyse the structure of texts to uncover the psychology of deception. These studies focus on linguistic aspects ranging from morphosyntax to semantic categories as well as discourse features and topic selection. Specifically, we use the LIWC software \cite{pennebaker2001linguistic} which is frequently used in text analysis to extract features reflecting certain psychological categories of interest \cite{robles2021jointly}. Each category has a corresponding dictionary of words, and the numerical feature value is computed as the percentage of words in the given text sample that are contained in the dictionary. Several available features were evaluated in \cite{chin2023written}, and our systems use 18 of those that were identified to significantly distinguish fake news from real news. See the details in Table \ref{tab:method:feature}.

\begin{table}
\renewcommand{\arraystretch}{1.3}
\centering
\caption{A list of the linguistic features used in this study with description and sample words. The features are highly related to fake news detection and selected based on linguistic expertise. Our \texttt{LingML} utilizes the features in different ways to enhance the system performance.}
\begin{tabular}{p{0.4\linewidth} p{0.5\linewidth}}
\hline\hline
\textbf{Feature} & \textbf{Description \& Examples} \\ \hline
feeling & feel, hard, cool, felt, etc.  \\ \hline
assent  &  yeah, yes, okay, ok, etc. \\ \hline
perception & in, out, up, there, etc. \\ \hline
discrepancy (discrep) & would, can, want, could, etc. \\ \hline
certitude & really, actually, of course, real, etc. \\ \hline
causation (cause) & how, because, make, why, etc. \\ \hline
words per sentence (wps) & average words per sentence  \\ \hline
space & in, out, up, there, etc. \\ \hline
auditory & sound, heard, hear, music, etc. \\ \hline
all-or-none (allnone) & no. of words, e.g., all, never, always \\ \hline
motion & go, come, went, came, etc. \\ \hline
negative tone (tone\_neg) & bad, wrong, too much, hate, etc. \\ \hline
swear words (swear) & shit, fuck, fucking, damn, etc. \\ \hline
positive tone (tone\_pos) & good, well, new, love, etc. \\ \hline
exclamation points (exclam) & no. of exclamation points \\ \hline
question mark (qmark) & no. of question marks \\ \hline
netspeak & unofficial words, e.g., :), u, lol, haha. \\ \hline
conversation & yeah, oh, yes, okay, etc. \\ 
\hline\hline
\end{tabular}
\label{tab:method:feature}
\end{table}

\subsection{\texttt{LingM}: Linguistic-Only ML} 
\label{sec:method:LingM}
The extracted linguistic features can be utilized in ML in two ways. We present the first one here. We train an ML model with linguistic features only and we call the model \texttt{LingM}. Intuitively, linguistic features shall be useful for fake news detection, and \texttt{LingM} based on those features can achieve a high level of detection accuracy, even if a non-advanced ML algorithm is used. We purposely choose several popular and relatively basic ML algorithms, like support vector machines (\textsc{Svm}), decision tree (\textsc{Dt}), random forest (\textsc{Rf}), \textsc{XGBoost}, and shallow neural network (\textsc{Nn}). These models are designed for numerical features and cannot process raw text data efficiently like NLP models. If \texttt{LingM} based on those basic algorithms is shown to be effective, it to some extent highlights the effectiveness of the numerical linguistic features for fake news detection. We perform normalization\footnote{We use z-score standardization here, i.e., for a feature $f$, a data point's feature value $x_f$ is transformed to $(x_f-\mu_f)/(\sigma_f+\epsilon)$, where $\mu_f$ is the sample mean of the feature values, $\sigma_f$ is the sample standard deviation and $\epsilon = 10^{-6}$ is used to zero-division error.} for all the 18 linguistic features used in \texttt{LingM} with the expectation that the features in a standardized scale contribute proportionally to \texttt{LingM}'s detection performance. The detailed algorithm configurations are presented in Section \ref{sec:exp} for experiments.

\subsection{\texttt{LingL}: Linguistic-Enhanced \texttt{LLM}}
\label{sec:method:LingL}
\begin{figure}
  \centering  \includegraphics[width=0.48\textwidth]{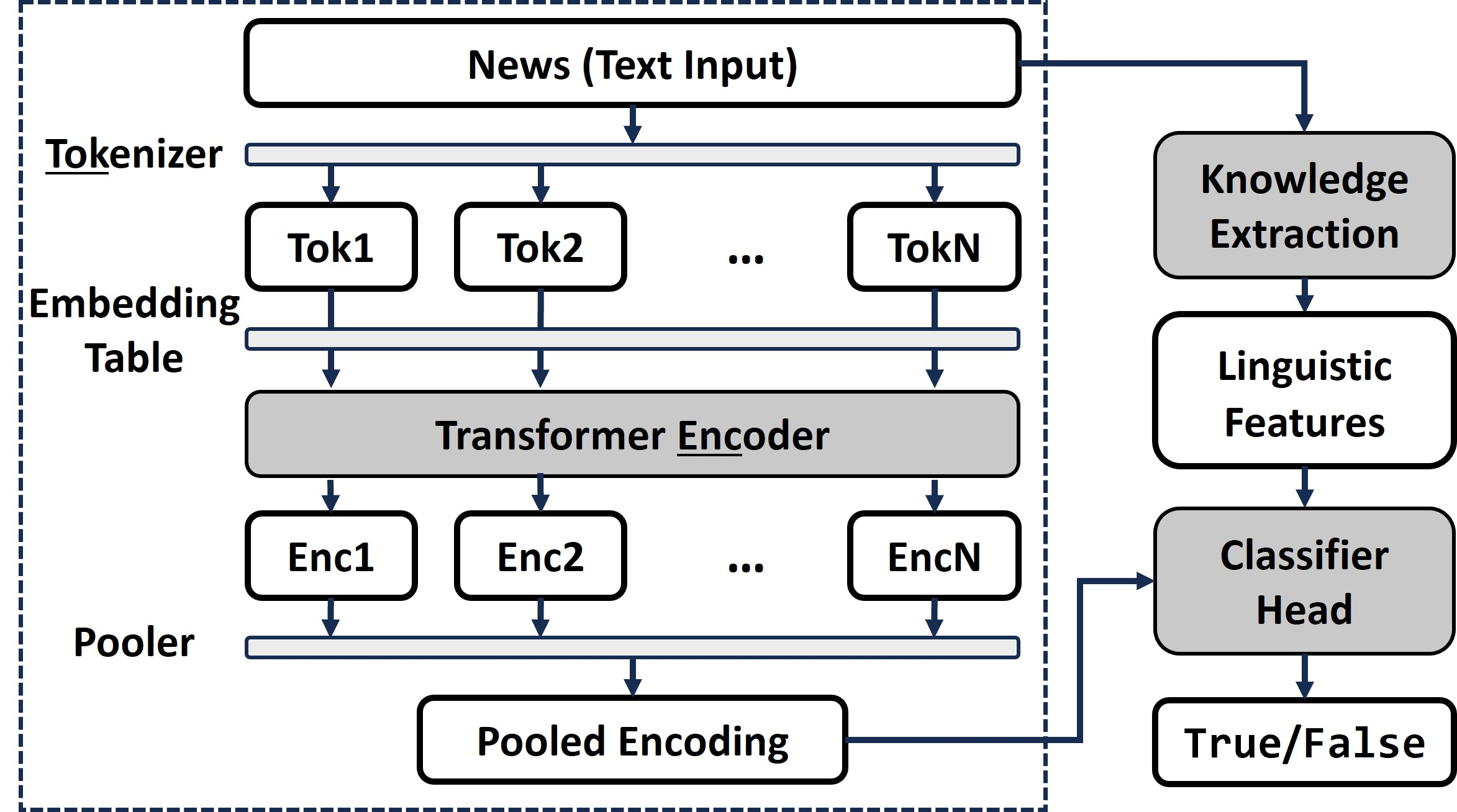}
\caption{An illustration of \texttt{LingL}'s system architecture with news as input. Linguistic knowledge is extracted from news and represented as numerical features, which are concatenated with the pooled encoding of the news produced by a \texttt{LLM}. The concatenated vector is directed to a classifier to produce the detection outcome of the news.}
\label{fig:LingL}
\end{figure}

We go beyond \texttt{LingM}'s basic usage of linguistic knowledge and propose \texttt{LingL} to incorporate hand-designed linguistic features into advanced transformer-based \texttt{LLM} \cite{vaswani2017attention}. Through this, we aim to combine the strength of scientific linguistics and \texttt{LLM}'s powerful text analysis capability for achieving optimal fake news detection performance. The fusion of linguistic knowledge and \texttt{LLM} is not trivial as most \texttt{LLM} are designed for text-based analysis with raw text (news) as the model's input. We design a fusion schema with two stages for \texttt{LingL} illustrated in Fig. \ref{fig:LingL}. In the first stage, we call a \texttt{LLM} to generate the encoding of the input text. For transformer models, this can be obtained by pooling the tokens' encodings. Essentially we use a \texttt{LLM} to map a text input to a numerical vector. We concatenate this encoding with the 18 linguistic features (after normalization) to obtain an enhanced vector representation of the input news. Then, we proceed to the second stage where this concatenated vector is processed by a multi-layered perceptron classifier which produces a \texttt{true} or a \texttt{false} label for the news input.

Existing \texttt{LLM} are mostly transformer-based models and BERT \cite{devlin2019bert} proposed in 2018 is one of the most representative ones for text analysis. In this paper, we consider BERT and several of its variants proposed afterwards (e.g., RoBERTa \cite{liu2019roberta} and CT-BERT \cite{muller2023covid}) that improve performance in different aspects. For each \texttt{LLM} in \texttt{LingL}, the weights and biases are initialized using the model checkpoints made available via Hugging Face. For models that are not specifically designed for text classification, the classifier head is randomly initialized following the strategy used for the base model. We present detailed configurations in \texttt{LingL}'s experimental setup. 

In summary, we extract linguistic knowledge as numerical features. We utilize the features as the input of an ML model for \texttt{LingM} and concatenate the features with \texttt{LLM}'s encoding for \texttt{LingL}. In the following section, we investigate the performance of \texttt{LingM} and \texttt{LingL} for fake news detection.
\section{Experimental Study} 
\label{sec:exp}
We present our experimental study in this section to demonstrate the impact of linguistics on ML-based fake news detection and we start with the data to be used in our experiments.

\subsection{Dataset and Data Pre-processing}
We describe the dataset for fake news detection and necessary data pre-processing as follows.

\subsubsection{Dataset Description}
Fake news and misinformation spread are among the daunting problems of the recent pandemic and the lessons we should learn from COVID-19. In certain social media groups, fake news even dominates the pandemic-related information with false claims about cures, prevention methods, origins of the virus, etc. This leads to confusion and even influences behavior contrary to recommended safety measures \cite{ahmed2022social}. In this paper, we consider a COVID-19 fake news dataset \cite{patwa2021fighting} published in an AAAI workshop as our case study. The dataset contains about six thousand news examples and is collected from several information sources including social media platforms (e.g., Facebook, Twitter, and Instagram), and official channels (e.g., public statements and press releases). Most fake news comes from social media, which however is also an important source of real news, e.g., verified Twitter accounts of the Centers for Disease Control (CDC) and the World Health Organisation (WHO). We follow the 60-20-20 split given in \cite{patwa2021fighting} where 60\%, 20\%, and 20\% of news are used for ML model training, validation during the training, and testing after the training, respectively. 

\subsubsection{Data Pre-processing}
\begin{figure}
\centering
\includegraphics[width=0.48\textwidth, trim={0 4.55in 0 0}, clip]{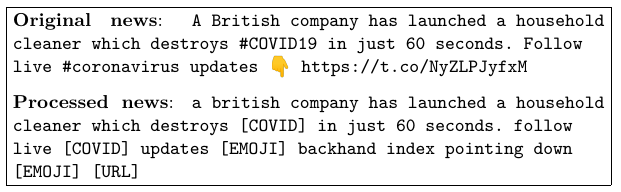}
\caption{A sample news from the dataset before and after data pre-processing with uncase, tokenization, and emoji conversion.}
\label{fig:news-sample}
\end{figure}

The raw dataset introduced above is not ready for ML analysis and we perform necessary data pre-processing for each news in the dataset. We have three main operations. The first one is uncasing, where all text is converted into lowercase. This removes unwanted biases that might have been generated due to the informal writing style on social media. The second operation is to generate new tokens mainly for certain special text components. For example, we replace URLs, mentions, hashtags, and COVID-related words with text-based tokens [URL], [MENTION], [HASHTAG], and [COVID], respectively. The last operation is for emoticons, which are widely used in social media. We replace each emoticon with its textual description obtained from the Python package \texttt{emoji} and the description is wrapped between two [EMOJI] tokens. A sample news before and after data pre-processing involving the above three operations is shown in Fig. \ref{fig:news-sample}. In addition, to further verify the reliability of the dataset, our linguist experts manually checked and updated the dataset label based on their domain knowledge.

\subsection{\texttt{LingM} Performance Analysis} 
\label{subsec:exp:LingM}
\texttt{LingML} models are trained with linguistic features only for fake news detection. We present our experimental study for \texttt{LingML} using the dataset introduced above.

\subsubsection{Experimental Setup}
We use Scikit-Learn to implement and train the small-scale \texttt{LingM} models \textsc{Svm}, \textsc{Dt}, \textsc{Rf}, and \textsc{XGBoost}. We search for the optimal hyper-parameters based on validation accuracy. To implement and train the \textsc{Nn}, we use the PyTorch framework. We configure 3 hidden layers each with 128 neurons with \texttt{GELU} activation. We use a dropout rate of 0.1 and \texttt{AdamW} optimizer for training. For each experiment, we perform five independent runs (e.g., training and testing) to account for randomness. We show the experimental results in Table \ref{tab:exp:LingM}.


\begin{table}
\renewcommand{\arraystretch}{1.3}
\centering
\caption{\texttt{LingM}'s performance (error rate) with five small-scale ML models trained with hand-designed linguistic features.}
\begin{tabular}{cccc}
\hline\hline
\multirow{2}{*}{\texttt{LingM}} & \multicolumn{3}{c}{Error Rate} \\ \cline{2-4} 
         & Mean     & Best     & SD   \\ \hline
\textsc{Svm} &   22.53\%   &  22.53\%  & 0.000 \\ \hline
\textsc{Dt} &   21.51\%   &  21.50\%  & 0.019 \\ \hline
\textsc{Rf}  &  19.11\% &   18.79\%   & 0.231 \\ \hline
\textsc{XGBoost} &   \textbf{18.32\%}   &  \textbf{18.32\%}  & 0.000 \\ \hline
\textsc{Nn} &   20.24\%   &  19.67\%  & 0.332 \\
\hline\hline
\end{tabular}
\label{tab:exp:LingM}
\end{table}

\subsubsection{Fake News Detection Performance}
Table \ref{tab:exp:LingM} demonstrates that linguistic knowledge is effective in differentiating fake news from real news. All five ML algorithms in our experiments are considered as basic in today's ML landscape with many large-scale and complex models. Yet, the error rate achieved by those small-scale algorithms with linguistic knowledge can be about 20\% on average, which is much better than random guess with half chances. Specifically, the mean error rate of \textsc{Svm}, \textsc{Dt}, \textsc{Rf}, \textsc{XGBoost}, and \textsc{Nn} are 22.5\%, 21.5\%, 19.1\%, 18.3\%, and 20.2\%, respectively. \textsc{XGBoost} outperforms the rest of the algorithms in terms of the mean and best detection performance. Linguistic knowledge lays a solid foundation for \texttt{LingM} to perform well, not only in terms of accuracy but also stability. The standard deviation (SD) of different runs is quite small for tested ML algorithms, e.g., 0.23 for \textsc{Rf} and infinitesimal for the best performing \textsc{XGBoost}. Overall, \texttt{LingM} reveals the potential of linguistic knowledge for fake news detection and the knowledge brings the benefits of accuracy and stability. 

\subsubsection{The Importance of Linguistic Knowledge}
Besides the aggregated impact of all features described above, we also plan to investigate the impact of different features in fine-granularity and we present the results in the following section. For the analysis, we work with the \textsc{XGBoost}-based \texttt{LingM} model which achieved the best performance as discussed above. For this model, each time a feature is used to split a branch of a booster, the two newly created branches become more accurate. We use the average gain in performance brought in by a given feature as a measure of its importance to the classifier. The summation of the importance values of all features is normalized to 100\% for easy interpretation, and the importance of each feature is between 0\% and 100\%.

\begin{figure}[t]
\centering
\includegraphics[width=0.48\textwidth]{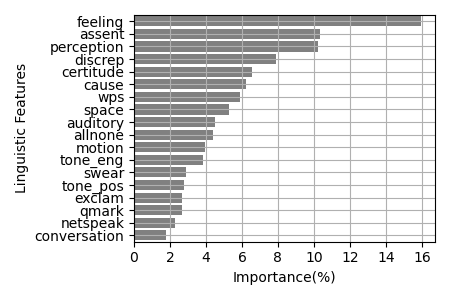}
\caption{The ranking of the features based on their importance (as percentage) to \texttt{LingM}. The summation of all importance/percentages is 100\%. Feature abbreviation is used for space saving and the detail can be found in Table \ref{tab:method:feature}.}
\label{fig:exp:feature}
\end{figure}

Fig. \ref{fig:exp:feature} shows each feature's importance for detecting fake news in \texttt{LingM}. We can see that the features do contribute to the detection output in different ways, with the importance ranging from 1.8\% to 15.9\%.  Among the features, \texttt{feeling} is the most important, with an importance of 15.9\% which is 3.6x of the median importance. Even compared to the second most important feature, \texttt{assent}, \texttt{feeling}'s importance is 54\% higher. We also notice that for the first five features (less than one-third of the total number) contribute to over 50\% of the importance. This observation highlights the opportunity for efficient usage of linguistic features by removing less important features to speed up the ML-based detection. In the following part, we consider the special case of detecting fake news using only the most important feature.

\subsubsection{Detection with One Linguistic Feature}
Some linguistic features are more relevant to fake news detection than others. Here, we investigate the possibility of detecting fake news with only one feature, and we choose the most important one, \texttt{feeling}, based on Fig. \ref{fig:exp:feature}. For maximum interpretability of the decision rule, we choose to model the input's veracity using the \textsc{Dt}-based \texttt{LingM}. In this case, the model can be simplified into an if-else logic reflected in the tree structure of \textsc{Dt}. We present the pseudo-code of the decision rule learnt by the \textsc{Dt} in Algorithm \ref{alg:exp:feeling}. Our experiments show that such (extremely) simplified \texttt{LingM} can achieve 28.4\% error rate. This is understandably higher than its \textsc{Dt} counterpart with all features but impressive with 6.9\% rate difference given that $\sim$95\% features are not utilized. The results demonstrate the strong correlation between linguistic knowledge and fake news detection and motivate enhanced linguistic knowledge integration with advanced \texttt{LLM}.

\addtolength{\topmargin}{0.01in}
\begin{algorithm}[t]
    \KwFeature{a news $t$}
    $x_1$ $\gets$ the value of the linguistic \texttt{feeling} feature of $t$\;
    \eIf{$x_1 <= -0.174$}{\textbf{return} \texttt{true};\hfill$\triangleright$ this is a fake news}{\textbf{return} \texttt{false};\hfill$\triangleright$ this is not a fake news}
\caption{\textsc{Dt} with with linguistic feeling feature } 
\label{alg:exp:feeling}
\end{algorithm}

\subsection{\texttt{LingL} Performance Analysis} 
\label{subsec:exp:LingL}
Neither \texttt{LingM} nor \texttt{LLM} utilizes the merits of both linguistics and ML well at the same time. For the former, basic ML algorithms learn from the linguistic data only. The latter explores the most advanced ML systems to process text data but without specific linguistic knowledge usage. We study the integration \texttt{LingL} here.

\subsubsection{Experimental Setup}
We perform an experimental study for our proposed \texttt{LingL} which integrates linguistic knowledge and \texttt{LLM}. We consider 11 transformer-based \texttt{LLM}, of which CT-BERT \cite{muller2023covid}, BERTweet \cite{nguyen2020bertweet}, and Twitter-RoBERTa \cite{barbieri2020tweeteval} are specialised for working with the informal language used on social media, while the others (e.g., ALBERT \cite{lan2020albert}, BERT \cite{devlin2019bert}, DistilBERT \cite{sanh2019distilbert}, Longformer \cite{beltagy2020longformer}, RoBERTa \cite{liu2019roberta}, XLM \cite{conneau2019cross}, XLM-RoBERTa \cite{conneau2020unsupervised}, and XLNet \cite{yang2019xlnet}) are trained on more formal data sources. For each tested \texttt{LLM}, we train the model for 10 epochs. We configure early-stopping for training to avoid over-fitting and validate the detection performance every 600 training steps. We also conduct five independent runs for each experiment and report the statistics.

\begin{table}
\renewcommand{\arraystretch}{1.3}
\centering
\caption{\texttt{LLM} and \texttt{LingL}'s classification error rates with different base models. The last column is calculated as the improvement in mean error rate. All values except SD are reported as percentages (\%).}
\begin{tabular}{@{\extracolsep{3pt}}cccccccc@{}}
\hline\hline
\multirow{2}{*}{Model Name} & \multicolumn{1}{c}{\texttt{LLM}} & \multicolumn{3}{c}{\texttt{LingL}} & \multirow{2}{*}{Improv.} \\ \cline{2-2} \cline{3-5} 
         & Mean     & Mean     & Best     & SD   & \\ \hline
ALBERT \cite{lan2020albert} & 3.37 & 2.99 & 2.80 & 0.11 & 11.4 \\ \hline
BERT \cite{devlin2019bert} & 3.21 & 2.75 & 2.57 & 0.10 & 14.5 \\ \hline
BERTweet \cite{nguyen2020bertweet} & 2.87 & 2.61 & 2.48 & 0.13 & 9.1 \\ \hline
CT-BERT \cite{muller2023covid} & 2.11 & \textbf{1.81} & \textbf{1.68} & 0.08 & 14.0 \\ \hline
DistilBERT \cite{sanh2019distilbert} & 3.08 & 2.78 & 2.66 & 0.06 & 10.0 \\ \hline
Longformer \cite{beltagy2020longformer} & 3.40 & 2.89 & 2.76 & 0.08 & 15.1 \\ \hline
RoBERTa \cite{liu2019roberta} & 3.20 & 2.80 & 2.38 & 0.22 & 12.3 \\ \hline
Twitter-RoBERTa \cite{barbieri2020tweeteval} & 2.94 & 2.72 & 2.52 & 0.11 & 7.6 \\ \hline
XLM \cite{conneau2019cross} & 3.30 & 2.71 & 2.48 & 0.14 & \textbf{17.9} \\ \hline
XLM-RoBERTa \cite{conneau2020unsupervised} & 2.87 & 2.57 & 2.38 & 0.12 & 10.4 \\ \hline
XLNet \cite{yang2019xlnet} & 3.36 & 3.09 & 2.94 & 0.12 & 7.8 \\ \hline
\hline
\end{tabular}
\label{tab:exp:LingL}
\end{table}

\subsubsection{\texttt{LLM} Performance}
We aim to analyze \texttt{LingL}'s performance and we start with the baseline \texttt{LLM} without linguistic knowledge usage. We report the results in Table \ref{tab:exp:LingL} and we have the following observations. \texttt{LLM} is effective for our application. All the tested models can push the detection error rate below 4\%. The best-performed \texttt{LLM} is CT-BERT \cite{muller2023covid}, which is reasonable as the \texttt{LLM} is specialized for COVID-related content. Note that our \texttt{LingM} introduced above achieves about 20\% error rate in most runs and \texttt{LLM}'s performance improvement is significant. The implication here is that linguistic knowledge's effectiveness in detecting fake news is not on par with advanced ML models like \texttt{LLM}. However, we would like to mention that ML relies substantially on training data. When the real-world data to be analyzed does not share much similarity with the training data, the ML performance may drop significantly. Linguistic knowledge offers scientific support to \texttt{LingML} and helps ML models to be less sensitive to specific data distributions and to be more generalizable in practice. 

\subsubsection{\texttt{LingL} Outperforms \texttt{LLM}}
\texttt{LLM} performs well and \texttt{LingL} is even better. Seen from Table \ref{tab:exp:LingL}, \texttt{LingL} outperforms \texttt{LLM} in all the settings. To better comprehend the performance differences, we report the error rate as well as \texttt{LingL}'s improvement to \texttt{LLM}. The improvement for XLM-based \texttt{LingL} is the most significant with a 17.9\% lower error rate and the average improvement is 11.8\% for all the tested \texttt{LLM}. Importantly, we observe that \texttt{LingL} based on small models like ALBERT with 12 million parameters can significantly outperform much larger \texttt{LLM} like BERT and RoBERTa with 110 and 125 million parameters, respectively, simply by incorporating linguistic features. 

The results of improvement demonstrate that linguistic knowledge is useful and can further improve the performance of \texttt{LLM} significantly, despite the fact that \texttt{LLM} is already highly accurate. Improvement from \texttt{LLM}'s competitive error rates like 4\%, even minimal, is challenging. Linguistic support consistently enhances fake news detection in each of the tested \texttt{LLM}. Furthermore, the CT-BERT-based \texttt{LingL} can push the error rate below 2\% which is competitive with the state-of-the-art methods for COVID-19 fake news detection. The results highlight the importance and potential of leveraging the linguistic knowledge in ML-based solutions. 

It is worth mentioning that ML without linguistic knowledge may produce improved detection performance eventually with newer and more advanced \texttt{LLM}. However such performance improvement is often at the cost of increased model size and demand on data which can be costly. \texttt{LingML} offers a different path to performance improvement without introducing significant overhead on model and data and improves interpretability and generalizability. For example, BERTweet was fine-tuned from its base model BERT with a 80GB corpus of 850M English tweets. Similarly, RoBERTa's domain-adapted version, Twitter-RoBERTa, used additional 58M tweets. Such fine-tuning is highly costly, and replacing such fine-tuning with our idea of incorporating linguistic features brings even better detection performance. This highlights the importance of leveraging the strength of linguistic knowledge to produce improved detection systems. 

\section{Conclusion and Future Works} \label{sec:conclusion}
In this paper, we proposed \texttt{LingML} to integrate linguistic knowledge into ML for fake news detection. Linguistic knowledge is extracted from the news with linguistic domain expertise and shown to be effective in discovering fake news-related patterns in ML. We specifically considered \texttt{LingM} by training ML models with linguistic features only and \texttt{LingL} by embedding linguistic features into advanced \texttt{LLM}. The experimental results demonstrated that linguistic knowledge is useful to improve detection performance and offer interpretability to ML-based solutions. We achieved the best fake news detection error rate of 1.8\% and outperformed the best-performed \texttt{LLM} by 17\%. By integrating linguistic and psychological features into ML models, our interdisciplinary study has a significant impact on fake news detection in social media. The proposed solutions also inspire further research in the domain of physics-informed ML.

In the future, we plan to enhance our method for linguistic integration with \texttt{LLM} to further improve the performance. We also plan to introduce additional linguistic knowledge and domain expertise into \texttt{LingML}. Expanding our experiments for more topics beyond COVID is also interesting and helps us gain confidence in the real-world deployment of our solutions with maximized impact.

\section*{Acknowledgement}
The authors acknowledge the contribution of the undergraduate students Brian Tan Jun Wei, Karissa Chua Hanyi, Lim Xian Mao Elson, Matthew Lim Yu De, and Tan Jia Ding from the Singapore Institute of Technology for data cleansing and preliminary study in their integrative team project in AY22/23.

\bibliographystyle{IEEEtran}
\bibliography{reference}
\balance 

\end{document}